# Swift progress for robots over complex terrain

[1] Chen Li, [2] Feifei Qian

[1] Department of Mechanical Engineering, Johns Hopkins University, Baltimore, Maryland 21218, USA; Terradynamics Lab: https://li.me.jhu.edu/

[2] Ming Hsieh Department of Electrical Engineering, University of Southern California, Los Angeles, California 90089, USA; RoboLAND Lab: https://sites.google.com/usc.edu/roboland

**A four-legged robot has learned to run on sand at faster pace than humans jog on solid ground. With low energy use and few failures, this rapid robot shows the value of combining data-driven learning with accurate yet simple models.**

Anyone who has walked along a sandy beach knows how hard it is to move on sand. Like other granular materials, including mud and snow[1], sand yields and flows under our feet until we sink deep enough, then it stops flowing and provides a stable foothold below the surface. Sand also doesn't spring back after we push it away, and the force it can support before giving way depends on how wet and how compact it is, thus changing how much our foot sinks in and slips as we walk[1]. These complexities complicate the task of controlling a robot so that it can run on sand. However, writing in *Science Robotics*, Choi *et al.*[2] have succeeded in doing so, enabling a four-legged robot to be fast, robust and energetically efficient on sand.

Legged robots have been able to run on solid ground for several decades[3–5], and some robots that are small enough to fit in the palm of your hand have even done so on uniform sand in the laboratory[6]. Larger legged robots can walk slowly on natural granular materials[7,8], but researchers have struggled to control legged robots such that they match an animal's running pace on sand. Choi *et al.* managed this feat — achieving a top running speed of 3 metres per second — by integrating three approaches.

First, they used reinforcement learning[8] to train their robot to maximize how fast it runs and minimize how often it fails and how much energy it expends. To do so, they applied a technique called privileged learning, which is like training a teacher so that they can teach a student efficiently[9]. A simulated robot — the teacher — first trains itself to identify optimal control strategies by learning from a very large dataset, which takes a long time. The student — the real robot — then benefits from what the teacher has already learned, and can use partial, noisy data to quickly shift between control strategies. In the authors' case, the teacher learned how to run across different sandy conditions in simulation, so that the student could adapt as it ran across real sand.

Second, to bridge the gap from simulation to reality, Choi *et al.* trained their 'teacher' by simulating sand with highly variable physical properties and load-bearing capacities, similar to those found in nature (wet to dry, loose to compact, Fig. 1). This is important because machine-vision systems, which are designed to see and interpret the world as we do with our eyes, cannot reliably estimate the physical properties of a challenging terrain. For example, it might erroneously classify the top layer of wet sand as dry. Because dry sand flows more easily than wet sand, this error in judgement will affect the robot's performance. By training the robot to learn to adapt to variable physical properties of sand, Choi *et al.* addressed this problem.

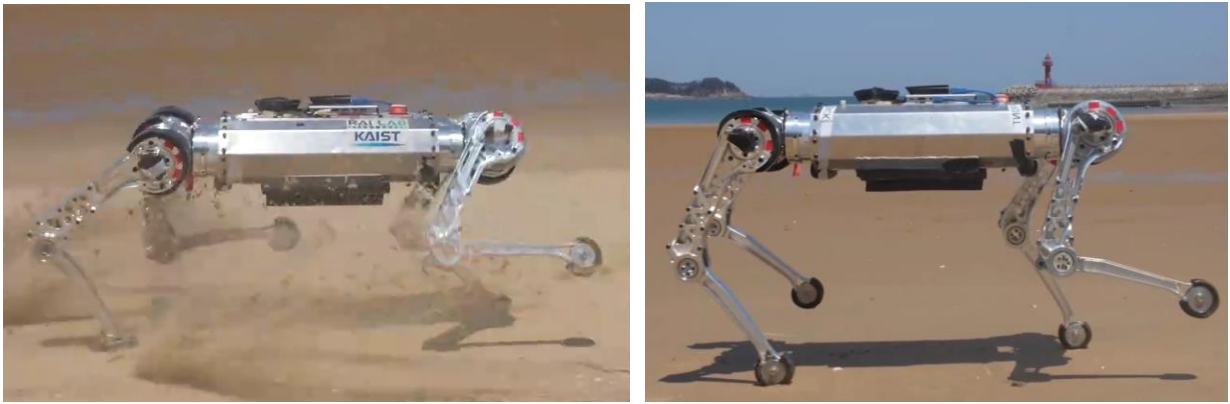

**Figure 1 | A robot learns to run on sand.** Choi et al.[2] used reinforcement learning to train a legged robot to run on level beach sand at a top speed of 3 metres per second. The robot's rapid, robust and efficient running was achieved by training it to adapt to variable terrain conditions (including dry and loosely packed sand, left, and wet and compact sand, right), leveraging the power of both data-driven learning and simulation approaches and models derived from fundamental research. (Adapted from Supplementary Movie S1 of ref. 2.)

Finally, to train their simulated robot, the authors selected and refined a model that describes the reaction forces exerted by sand on their robot's small feet as they impact sand[10]. This model was not only accurate enough to capture the interaction physics, but also simple enough to make their simulations fast, both of which were necessary for the authors' success.

Aside from their robot's remarkable performance and robustness, Choi and colleagues' work is also notable in how well it integrated machine learning with models[11]. As more and more engineers aspire to achieve ever higher robot mobility in the real world using machine learning and simulations, these authors still appreciate the value of models developed through rigorous experimental research. They clearly went to great lengths to review the growing literature on the forces exerted by objects moving in sand, much of which comes from experimental physics research involving animal and robot locomotion in the laboratory[12].

This knowledge enabled Choi et al. to understand the nuanced contributions to forces on a small foot as it rapidly impacts sand[10]. One such contribution comes from the pressure exerted by the weight and friction of the sand particles. Another is a dynamic contribution due to particle inertia, which is similar to aerodynamic or hydrodynamic drag felt in fluids. A third factor — transient yet large — also comes from particle inertia, resulting from the sudden acceleration of the sand particles when they are first impacted by the foot. The authors demonstrated that their refined model (which includes all these force contributions) allowed the robot to achieve much higher performance and robustness than is possible with less accurate (and still often used) models.

Future research is still needed to improve legged robots so that they have animal-like mobility in terrain that is even more challenging than beach sand. As Choi and colleagues' robot moved on beach sand, only its feet sank into sand — a scenario that is well described by the simple model that the authors used. However, achieving such success would be much more challenging if the robot's long legs were to sink more deeply into sand[13,14] — for example, if it were carrying a person or a large package. The robot would also face difficulties if it kept stepping into sand that was already disturbed[13], because this sand would continuously change its load-bearing properties. It would similarly struggle to move on sand dunes, which avalanche when disturbed[15]. Enabling robots to learn to deal with these extreme situations requires that we train them using accurate yet simple models that capture these additional complex behaviours of sand, such as that in ref. 1 and 16 or models yet to be developed.

Choi and colleagues' demonstration is especially valuable and timely, given the current trend of data-driven learning approaches becoming pervasive. These techniques have been successful for solving problems such as image classification, medical diagnosis, natural language generation and game playing. But the authors' work is a reminder that good models are just as essential as data-driven learning for tackling problems like robot locomotion in complex terrains. In many-particle systems such as sand, phenomena emerge that cannot easily be reconstructed from the fundamental laws of nature[16]. This calls for basic experimental research into the underlying principles and mechanisms, which machine learning might miss[11], and which simulations that are not vigorously validated by experiments might not fully capture[12].

Aerial and underwater vehicles that are autonomous, safe, fast and efficient have been engineered successfully because we have a fundamental understanding of aerodynamics and hydrodynamics. Choi *et al.* have set an excellent example of how the same level of success can be achieved for animal-like robots traversing natural terrains, building on the foundation of the emerging field of terradynamics[1]. By integrating fundamental research with data-driven learning and simulation approaches[11] in this way, we foresee exciting and rapid progress towards the goal of developing robots that can move across all terrains.